\PassOptionsToPackage{x11names}{xcolor}

\documentclass[10pt, a4paper]{article}

\usepackage{misc/lrec-coling2024} %

\usepackage{times}
\usepackage{latexsym}
\usepackage{multicol}
\usepackage{svg}
\usepackage[T1]{fontenc}

\usepackage[utf8]{inputenc}

\usepackage{microtype}

\usepackage{inconsolata}

\usepackage{CJKutf8}
\usepackage{makecell}
\usepackage{multirow}
\usepackage{arydshln}

\usepackage[most]{tcolorbox}
\usepackage{todonotes}
\usepackage[x11names]{xcolor}
\usepackage{mdframed}
\usepackage{amssymb, amsmath}
\usepackage{cleveref}
\usepackage{booktabs}
\usepackage{algorithm, algorithmicx, algpseudocode}
\usepackage{scalerel} %
\usepackage{enumitem} %
\usepackage{amsthm} %
\usepackage{thmtools} %
\usepackage{thm-restate} %
\usepackage{array}
\usepackage{caption}
\usepackage{xspace}
\usepackage{soul}

\usepackage{xfrac}

\usepackage{arevmath}

\crefname{myexample}{Example}{}
\DeclareCaptionType{myexamplef}[Example][List of Examples]
\crefname{myexamplef}{Example}{}
\crefname{prooff}{Proof}{}
\DeclareCaptionType{prooff}[Proof][List of Proofs]

\definecolor{TodoColor}{rgb}{1,0.7,0.6}

\newmdenv[
  linecolor=black,
  linewidth=1.2pt,
  topline=false,
  bottomline=false,
  rightline=false,
  innertopmargin=0mm,
  innerbottommargin=-0.5mm,
  innerleftmargin=1mm,
  skipabove=1.1\topsep,
  skipbelow=0.5\topsep,
]{quotebox}

\newcommand{\hrefEmail}[2]{\texttt{\href{mailto:#1}{\color{black}{\fontsize{11}{9}\selectfont #2}}}}
\newcommand{\proofpreamble}{\fontsize{8.2}{2}\selectfont}
\newenvironment{equationhalf}[1]
{\begin{minipage}{#1\linewidth}
\begin{equation}
}
{
\end{equation}
\end{minipage}
}

\newcommand{\merge}{\mu}
\newcommand{\merges}{\boldsymbol{\merge}}
\newcommand{\mop}[2]{\ensuremath{[#1,#2]}}
\newcommand{\mopt}[2]{\ensuremath{[\texttt{#1},\texttt{#2}]}}

\newcommand{\vocab}{\ensuremath{\mathcal{V}}}

\newcommand{\dec}{\scalebox{0.9}{\textsc{Dec}}}
\newcommand{\sub}{\scalebox{0.9}{\textsc{Sub}}}

\newcommand{\ent}{\mathrm{H}}
\newcommand{\eff}{\mathrm{Eff}}
\newcommand{\pdelta}{p_{\scaleto{\vocab}{4pt}}}
\newcommand{\rvDelta}{\mathrm{W}_{\scaleto{\vocab}{4pt}}}
\newcommand{\cv}{c_{\scaleto{\vocab}{4pt}}}
\newcommand{\cvp}{c_{\scaleto{\ensuremath{\mathcal{V}'}}{4pt}}}
\newcommand{\pv}{p_{\scaleto{\vocab}{4pt}}}
\newcommand{\pvp}{p_{\scaleto{\ensuremath{\mathcal{V}'}}{4pt}}}
\newcommand{\Tv}{T_{\scaleto{\vocab}{4pt}}}
\newcommand{\Tvp}{T_{\scaleto{\ensuremath{\mathcal{V}'}}{4pt}}}

\newcommand{\prooftext}[1]{\textit{\textcolor{gray}{#1}}}

\let\svthefootnote\thefootnote
\newcommand\blankfootnote[1]{%
  \let\thefootnote\relax\footnotetext{#1}%
  \let\thefootnote\svthefootnote%
}

\definecolor{ethblue}{rgb}{0,0.1,0.4}
\newcommand{\ethletter}{\hspace{-0.5mm}\text{
    \fontfamily{phv}\fontseries{bx}\fontsize{8}{\baselineskip}\selectfont
    \textit{\textbf{\color{ethblue}{E}}}}
}
\newcommand{\tokyoletter}{
    \text{\raisebox{-0.3mm}{\includegraphics[width=3mm]{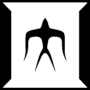}}}
} 
\title{Two Counterexamples to \textit{Tokenization and the Noiseless Channel}}

\name{Marco Cognetta$^{\tokyoletter}$, Vilém Zouhar$^{\ethletter}$, Sangwhan Moon$^{\tokyoletter}$, Naoaki Okazaki$^{\tokyoletter}$}

\address{$^{\tokyoletter}$Tokyo Institute of Technology, School of Computing\\ $^{\ethletter}$ETH Zurich, Department of Computer Science \\
         \hrefEmail{cognetta.marco@gmail.com}{cognetta.marco@gmail.com}, \hrefEmail{vzouhar@ethz.ch}{vzouhar@ethz.ch}, \hrefEmail{sangwhan@iki.fi}{sangwhan@iki.fi}, \hrefEmail{okazaki@c.titech.ac.jp}{okazaki@c.titech.ac.jp}}

\abstract{
In \textit{Tokenization and the Noiseless Channel} \cite{zouhar-etal-2023-tokenization}, {Rényi efficiency} is suggested as an intrinsic mechanism for evaluating a tokenizer: for NLP tasks, the tokenizer which leads to the highest Rényi efficiency of the unigram distribution should be chosen.
The Rényi efficiency is thus treated as a predictor of downstream performance (e.g., predicting BLEU for a machine translation task), without the expensive step of training multiple models with different tokenizers.
Although useful, the predictive power of this metric is not perfect, and the authors note there are additional qualities of a good tokenization scheme that Rényi efficiency alone cannot capture.
\newline We describe two variants of BPE tokenization which can arbitrarily increase Rényi efficiency while decreasing the downstream model performance.
These counterexamples expose cases where Rényi efficiency fails as an intrinsic tokenization metric and thus give insight for building more accurate predictors.
 \\ \newline  }

\begin{document}

\maketitleabstract

\section{Introduction}

\begin{figure}[htbp]
\includegraphics[width=\linewidth]{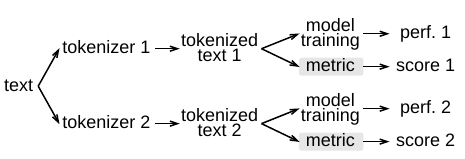}
\caption{The same model is trained on two different tokenizations. We are looking for \graybox{metric} such that, across multiple settings, \textit{score 1 > score 2} if and only if \textit{performance 1 > performance 2}.}
\label{fig:tokenization_diagram}
\vspace{-2mm}
\end{figure}

The tokenizer choice strongly impacts the downstream NLP model performance \citep{domingo2019much}.
At the same time, it is difficult to select the best one, because comparing two tokenizers typically requires fully training a model on top of each tokenizer, which can be prohibitively expensive, taking days or weeks of compute times.
This motivates the search for an intrinsic evaluation of tokenizers --- finding easy-to-compute metrics that can be evaluated using only the tokenized text and which signal if the tokenization will be good for the downstream task (see \Cref{fig:tokenization_diagram}).

In \textit{Tokenization and the Noiseless Channel}, \citet{zouhar-etal-2023-tokenization} propose a new metric based on a generalization of Shannon Entropy --- Rényi efficiency of the unigram distribution.
This quantity is slightly better correlated to downstream model performance (BLEU score on a translation task) than a percentile frequency metric, which they adopted based on \citet{gowda2020finding}.
It is also much better correlated than Shannon Entropy \cite{shannon1948mathematical} and average sequence length.
The authors suggest a characteristic of good tokenizers: not only should a tokenizer contain few rare tokens (tokens that appear infrequently in tokenized texts), but they should also not contain many very high-frequency tokens. 
While they propose this metric for intrinsically evaluating tokenizers before training, they note that it does not fully account for the relative performances of tokenizers and leave the search for other metrics as an open problem.
This means there is some variance in the performances of all possible tokenizers which the Rényi efficiency does not capture.

\definecolor{GrayMark}{rgb}{0.9,0.9,0.9}
\newtcbox{\graybox}{
    on line,
    colback=GrayMark, colframe=GrayMark, boxrule=0pt, arc=1mm, boxsep=-1mm, left*=1mm, right*=1mm,
}

In this work, we introduce two variants of BPE tokenizers for which we can explicitly increase Rényi efficiency while degrading the downstream model performance.
This goes against the aforementioned hypothesis where high Rényi efficiency implies better expected performance.
We hope that these examples will help spur more research into where Rényi efficiency and other related metrics accurately predict performance and where they fail.
Ultimately, these efforts are intended to create a better predictor of downstream performance for NLP tasks based on the tokenized text.

\medskip
\noindent
\textbf{Structure:}
\begin{itemize}[noitemsep,topsep=0mm,left=3mm]
\item[\ref{sec:bpe})] Introduction to Byte-Pair Encoding.
\item[\ref{sec:renyi_efficiency})] Rényi efficiency, other metrics and their respective tokenization hypotheses.
\item[\ref{sec:counterexamples})] Two families of counterexamples to the Rényi efficiency tokenization hypothesis and proofs.
\item[\ref{sec:experiments})] Empirical verification of counterexamples with machine translation performance.\footnote{Code to reproduce all experiments will be made available.}
\end{itemize}

\section{Byte-Pair Encoding}
\label{sec:bpe}
Byte-Pair Encoding was proposed as a compression algorithm \citep{Gage1994ANA}, repurposed in NLP for machine translation \citep{sennrich2016neural}, and later formalized and bounded mathematically \citep{zouhar2023formal}.
The BPE algorithm has two parts: (1) training, where the merges are constructed, and (2) tokenization, where the merges are applied to new text.
In this paper, we focus only on the \textit{tokenization} side and we use the standard BPE \textit{training} algorithm for all experiments.

\paragraph{BPE Training.}
A trained BPE tokenizer is a triplet $(\Sigma, \vocab, \merges)$, where $\Sigma$ is a set of \textit{atomic} characters, $\vocab \subset \Sigma^+$ is the \textit{subword vocabulary}, and $\merges \in \vocab{\times}\vocab$ is an ordered list of \textit{merges}.
The $\merges$ defines how subwords combine to form new subwords.
We use the notation of
$\textsc{Apply}(\mop{a}{b}, t)$ to \textit{apply} a merge to a sequence -- that is, to replace each instance of consequent symbols $a b$ with $ab$ in the sequence.

See examples of notation:

\vspace{1mm}
\noindent
\resizebox{\linewidth}{!}{
\renewcommand{\arraystretch}{1.4}
\begin{tabular}{l<{\hspace{-4mm}}l}
\toprule \\[-2.2em]
$\Sigma$ & \ttfamily {a, b, c, d, $\ldots$} \\[-0.2em]
$\vocab$ & \ttfamily {a, the, cow, token, -ization, $\ldots$} \\
$\merges$ & \ttfamily \mopt{t}{h}, \mopt{\mopt{t}{h}}{e}, $\ldots$ \\
$\textsc{Apply}$ & $(\mopt{c}{-o}, \texttt{c -o -w}) \rightarrow \texttt{co -w}$, $\ldots$ \\
\bottomrule
\end{tabular}
}
\medskip

We construct a BPE tokenizer using the standard algorithm described by \citet{Gage1994ANA,sennrich2016neural,zouhar2023formal}.
In our notation, $\merges$ is ordered by the order in which a merge was added to the BPE vocabulary during construction.
The first element of $\merges$ is the first pair that merged during BPE initialization, and we proceed in decreasing order from there.
To see how the merge sequence $\merges$ is constructed.
For completeness, the BPE \textit{training} algorithm will be included in the Appendix.

\paragraph{BPE Tokenization.}
We describe the standard BPE tokenization procedure in \Cref{algo:bpe_inference}, which serves as the baseline.
The BPE tokenization algorithm is applied independently on all words in the corpus, and so \Cref{algo:bpe_inference} describes an algorithm that takes a single word (as a sequence of characters from $\Sigma$) as input along with the BPE tokenizer.
We modify this algorithm and its inputs for our experiments later.
In words, the BPE tokenization algorithm starts with a sequence of characters and iteratively merges the first pair in $\merges$, then the second and ultimately the last.

\begin{algorithm}[h]
{\fontsize{11}{12}\selectfont
\begin{algorithmic}[1]
\State $t \gets \Call{CharacterSequence}{w}$
\For{$\mop{a}{b} \in \merges$}
    \State $t \gets \Call{Apply}{\mop{a}{b}, t}$
\EndFor
\State \Return $t$\label{algo:inference_return}
\end{algorithmic}
}
\caption{%
    \hspace{-1mm}: Standard BPE tokenization\newline
    \textbf{Inputs}: word $w \in \Sigma^*$, merges $\merges$,\newline
    \text{}\qquad\qquad$\Sigma \cup \bigcup_{\mop{a}{b} \in \merges} \{ab\} = \vocab$\newline
    \textbf{Output}: tokenized sequence $t \in \vocab^*$
}
\label{algo:bpe_inference}
\end{algorithm}

\vspace{-3mm}
\section{Rényi Efficiency and Other Metrics}
\label{sec:renyi_efficiency}

The performance-from-tokenization prediction task can be formalized as finding a scalar metric $m$ over a sequence of tokens from $\vocab^+$, therefore $m: \vocab^+ \rightarrow \mathbb{R}$.
The goal for this metric is to correlate with some downstream performance, such as BLEU of a machine translation system.
\vspace{-5mm}

\paragraph{Baseline.}
The simplest metric, denoted $C$, is to use the average number of tokens per line.
The hypothesis is that given the same vocabulary size, if the tokenized text contains fewer tokens, then the tokenization is likely better.
In the following example, the first tokenization is likely better because the text is tokenized into fewer pieces.
\medskip

\noindent
\resizebox{\linewidth}{!}{
\renewcommand{\arraystretch}{1.2}
\begin{tabular}{l}
\toprule
$\mathrm{C}($\texttt{the quick fox jump -ed}$)=5$ \\
$\mathrm{C}($\texttt{th -e br -own fox j -u -m -ped}$)=9$ \\
\bottomrule
\end{tabular}
}
\medskip

With a large enough vocabulary size, one could represent each word as a separate token.
This tokenization would obtain the lowest possible score by $\mathrm{C}$ and be marked as the best one, even though it would yield very poor downstream task results because of the lack of generalization to unseen words \citep{sennrich2016neural}.

\paragraph{Percentile Frequency.}
For more nuanced metrics, we turn to the unigram distribution of the tokenized text, $\pdelta$.
For example $\pdelta(\text{``the''})=0.02$.
Given the start and end percentiles $\gamma_1$ and $\gamma_2$, the quantity is defined as:
\begin{align}
F_{\gamma_1,\gamma_2}(\vocab) = \sum_{w\, \in\, \textsc{Perc.}_{\pdelta}(\gamma_1, \gamma_2)} \hspace{-6mm} \pdelta(w),
\end{align} and the higher the sum, the better the tokenizer should be \cite{gowda2020finding}. We use $F_{0.03,0.83}$ based on \cite{zouhar-etal-2023-tokenization}, which performed a hyperparameter scan to determine the percentiles which best correlated with downstream performance.

\paragraph{Rényi Efficiency.}
Finally, we bring to attention the Rényi entropy of the random variable $\rvDelta$ distributed according to $\pdelta$ and its associated quantity, Rényi efficiency:
\begin{align}
& \hspace{-1mm}\ent_\alpha(\rvDelta) =\! \lim_{\alpha'\rightarrow \alpha} \frac{1}{1-\alpha'}\log \left( \sum_{w \in \vocab} \pdelta(w)^{\alpha'} \right) \label{eq:Rényi_entropy} \\
& \hspace{-1mm}\eff_\alpha(\rvDelta) \approxeq \frac{\ent_{\alpha}(\rvDelta)}{\log |\vocab|}  \label{eq:Rényi_efficiency}
\end{align}

The higher the Rényi entropy, the more ``balanced'' the unigram distribution is, meaning there is a less pronounced difference between the frequencies of the most common and least common tokens.
We use the Rényi efficiency instead of Rényi entropy to normalize the effect of vocabulary size.
This works similarly with Shannon entropy and Shannon efficiency (when $\alpha = 1$).
This metric is found by \citet{zouhar-etal-2023-tokenization} to correlate the highest with the downstream performance on a translation task.

\section{Counterexamples}
\label{sec:counterexamples}

The Rényi Efficiency tokenization hypothesis states that the higher the Rényi efficiency of the unigram distribution, the better the downstream model performance.
To provide counterexamples, we want to find tokenizers that increase the Rényi efficiency while 
worsening the performance.
We provide two such counterexamples: \textsc{Random-Drop BPE} and \textsc{Duplicate BPE}.
In the Appendix we will also include a naïve method which can arbitrarily \textit{lower} the Rényi efficiency of a tokenizer (by inflating $\vocab$) while not affecting the performance, thus also serving as a counterexample.

\subsection{\textsc{Random-Drop BPE}}

Given a BPE tokenizer $(\Sigma, \vocab, \merges)$, we form a new tokenizer by selecting integer hyperparameters $1 \le k \le N \le |\vocab|$.
Then, we randomly select $k$ subwords from the top $N$ most frequent subwords in $\vocab$ (restricted to non-atomic subwords) to form a set $\mathcal{D}$.
Setting $N = |\vocab|$ would allow any non-atomic token to be a candidate for inclusion in $\mathcal{D}$.
A \textsc{Random-Drop BPE} tokenizer is then defined as $(\Sigma, \vocab, \merges, \mathcal{D})$.
The elements of $\mathcal{D}$ are marked for decomposition via a function $\dec$.
We omit the parameterization of $\dec$ on $(\Sigma, \vocab, \merges, \mathcal{D})$ for brevity.
\begin{align}
\raisebox{4mm}{\hspace*{-10mm}$\dec(z){=}\begin{cases} 
    \dec(x)\,\,\, \dec(y), & \hspace{-2mm}z = \mop{x}{y} \wedge z \in \mathcal{D}   \\
    z, & \hspace{-2mm}\text{otherwise}
\end{cases}$}\hspace{-10mm}&
\end{align}

\begin{figure}[htbp]
\centering
\includegraphics[width=0.95\linewidth]{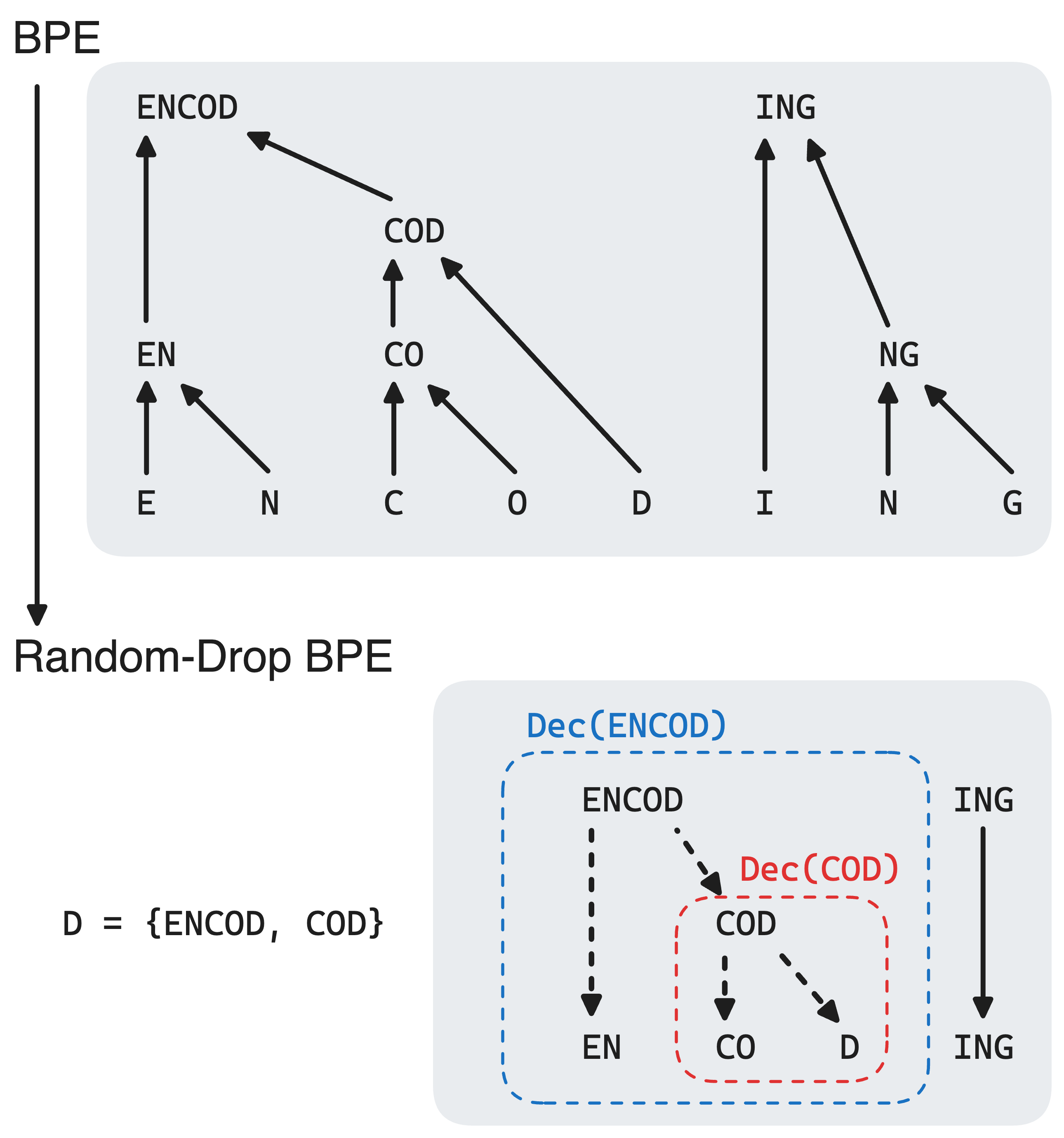}
\vspace{-1mm}
\caption{An example of \textsc{Random-Drop BPE} with $\mathcal{D} = \{\texttt{ENCOD},~\texttt{COD}\}$. The tokenization process is two-staged: (1) a regular BPE tokenizer is used to produce a tokenization, and (2) \textsc{Random-Drop BPE} recursively decomposes subwords that appear in the set $\mathcal{D}$ (first $\texttt{ENCOD}$, and then $\texttt{COD}$). Since $\texttt{ING}$ is not in $\mathcal{D}$, it is unchanged during the decomposition step.
The final tokenized text is \texttt{EN -CO -D -ING}.
If $\mathcal{D}$ was just $\{\texttt{ENCOD}\}$, then the final tokenization would have been \texttt{EN -COD -ING}, and if it was just $\{\texttt{COD}\}$, the final tokenization would have been $\texttt{ENCOD -ING}$, since $\texttt{-COD}$ was not in the final tokenization produced by the initial BPE pass.
}
\label{fig:random-drop}
\vspace{-3mm}
\end{figure}

Given a subword $z \in \mathcal{D}$, $\dec$ recursively undoes merges, splitting tokens into their merge pairs, until none of the tokens remaining in the sequence are in $\mathcal{D}$.
This decomposition is applied \textit{after} the regular BPE tokenization is complete.
See an example of the two-stage process in \Cref{fig:random-drop}.
The \textsc{Random-Drop BPE} tokenizer inference is implemented by replacing Line \ref{algo:inference_return} of \Cref{algo:bpe_inference} with:
\begin{align}
\textbf{return} \hspace{2mm} \langle \dec(t_i) \mid 1 \leq i \leq |t| \rangle
\end{align}

This tokenizer modification is distinct from BPE-Dropout regularization  \cite{provilkov-etal-2020-bpe}.
\textsc{Random-Drop BPE} is entirely deterministic (the set $\mathcal{D}$ is selected after vocabulary construction and is never changed), and so tokenizing a sequence multiple times will always result in the same sequence.
Also, the decomposition happens after the standard BPE tokenization.
In contrast, BPE-Dropout is performed during tokenization at each iteration of the merging algorithm by selectively blocking merges.
Given the same input, BPE-Dropout can produce different final tokenizations.

\paragraph{Efficiency Increase.}
In all of our experiments (\Cref{sec:experiments}), the \textsc{Random-Drop BPE} tokenizer improves Rényi efficiency compared to its underlying BPE tokenizer.
Now, we give some intuition for why this should happen empirically --- specifically, in the form of a proof of the sufficient conditions for a token to, if dropped, increase the Rényi \textit{entropy}.
The condition does not rely on the vocabulary size, and so does not relate directly to Rényi \textit{efficiency}, which introduces a divisor of $\log(|\vocab|)$.
Under the assumption that all subwords in the vocabulary appear at least once in the final tokenized corpus,\footnote{A subword can be a valid merge but not appear in any final tokenization if it always appears as a part of a longer subword. In this case, \textsc{Random-Drop BPE} can increase the vocabulary size if a specific drop is performed where both component subwords have 0 frequency. If the increase in entropy is too small, the larger normalizing factor can decrease the efficiency.}
increasing Rényi entropy always increases Rényi efficiency, because we are effectively removing one token from the vocabulary and the normalizing factor is reduced to $\log(|\vocab| - 1)$.

\begin{restatable}{theorem}{thmrandomdropbpe}
\label{thm:randomdropbpe}
Let $c_{\vocab}(w)$ be the unigram frequency of $w{\in} \vocab$, and $T_{\vocab} = \sum_{w \in \vocab} c(w)$ so that $p_{\vocab}(w) = \frac{c_{\vocab}(w)}{T_{\vocab}}$ is the unigram distribution. Suppose subword $x = (y, z) \in \vocab$ is chosen for decomposition, resulting in a new vocabulary $\vocab' = \vocab \backslash \{x\}$. Let $W_{\vocab}$ and $W_{\vocab'}$ be the corresponding random variables and let $\alpha > 1$ be the parameter for Rényi entropy. Then $\ent_\alpha[W_{\vocab}] < \ent_\alpha[W_{\vocab'}]$ if 
\begin{align*}
& \left(1 + \frac{c_{\vocab}(x)}{T_{\vocab}} \right)^\alpha > 1 \\
&\hspace{10mm} + \large( (c_{\vocab}(x) + c_{\vocab}(y))^\alpha + (c_{\vocab}(x) + c_{\vocab}(z))^\alpha \\
& - c_{\vocab}(x)^\alpha - c_{\vocab}(y)^\alpha - c_{\vocab}(z)^\alpha \large)/\left(\Sigma_{w \in \vocab} c_{\vocab}(w)^\alpha\right)
\end{align*}
\end{restatable}

\begin{figure*}[htbp]
\proofpreamble

Let $\alpha > 1$. When removing $x$ from the vocabulary, every instance of $x$ in the tokenized sequence gets replaced with $(y, z)$. Thus, in $\vocab'$, $y$ and $z$'s counts are increased by $\cv(x)$ and the overall total number of tokens is increased by $\cv(x)$ (as the total change is $2\cv(x) - \cv(x)$), and so $\Tvp = \Tv + \cv(x)$, $\cvp(x) = 0$, $\cvp(y) = \cv(x) + \cv(x)$, $\cvp(z) = \cv(x) + \cv(z)$, and $\cvp(w) = \cv(w)$ for all other $w \in \vocab \backslash \{x, y, z\}$. Then:

\begin{equationhalf}{0.35}
\ent_\alpha[\rvDelta] < \ent_\alpha[\rvDelta']
\end{equationhalf}
\begin{equationhalf}{0.64}
\frac{1}{1-\alpha}\log \left( \sum_{w \in \vocab} \pv(w)^{\alpha} \right)
< \frac{1}{1-\alpha}\log \left( \sum_{w \in \vocab}\pvp(w)^\alpha \right)
\end{equationhalf}

\begin{equationhalf}{0.35}
\sum_{w \in \vocab} \pv(w)^{\alpha}
> \sum_{w \in \vocab}\pvp(w)^\alpha 
\end{equationhalf}
\begin{equationhalf}{0.64}
\frac{ \sum_{w \in \vocab} \cv(w)^\alpha}{\Tv^\alpha} > \frac{\sum_{w \in \vocab'} \cvp(w)^\alpha}{\Tvp^\alpha} 
\end{equationhalf}

\begin{equationhalf}{0.35}
\left( \frac{\Tvp}{\Tv} \right)^\alpha > \frac{\sum_{w \in \vocab'} \cvp(w)^\alpha}{\sum_{w \in \vocab} \cv(w)^\alpha}
\end{equationhalf}
\begin{equationhalf}{0.64}
\left( \frac{\Tvp}{\Tv} \right)^\alpha = \left(\frac{\Tv + \cv(x)}{\Tv} \right)^\alpha = \left(1 + \frac{\cv(x)}{\Tv} \right)^\alpha  > \frac{\sum_{w \in \vocab'} \cvp(w)^\alpha}{\sum_{w \in \vocab} \cv(w)^\alpha} \label{eq:finalrighthand} 
\end{equationhalf}

\medskip
In \Cref{eq:finalrighthand} the left-hand side is in the form we want. The right-hand-side numerator, $\sum_{w \in \vocab'} \cvp(w)^\alpha$, can be rewritten as:
\begin{align}
\sum_{w \in \vocab'} \cvp(w)^\alpha = \Bigg( \hspace{-1mm}\sum_{\substack{w \in \vocab\\ \backslash \{x, y, z\}}}\hspace{-3mm} \cv(w)^\alpha \Bigg) {+} \cvp(y)^\alpha {+} \cvp(z)^\alpha = \Bigg( \sum_{w \in \vocab} \cv(w)^\alpha \Bigg) {+} \cvp(y)^\alpha {+} \cvp(z)^\alpha {-} \cv(x)^\alpha - \cv(y)^\alpha {-} \cv(z)^\alpha
\end{align}
Noting that $\cvp(y) = \cv(x) + \cv(y)$ and $\cvp(z) = \cv(x) + \cv(z)$ and that $\sum_{w \in \vocab} \cv(w)^\alpha$ is the right-hand-side denominator of \Cref{eq:finalrighthand}. We arrive at our final right-hand side:

\begin{align}
\frac{\sum_{w \in \vocab'} \cvp(w)^\alpha}{\sum_{w \in \vocab} \cv(w)^\alpha} &= 1 + \frac{\cvp(y)^\alpha + \cvp(z)^\alpha - \cv(x)^\alpha - \cv(y)^\alpha - \cv(z)^\alpha}{\sum_{w \in \vocab} \cv(w)^\alpha} \\
&= 1 + \frac{(\cv(x) + \cv(y))^\alpha + (\cv(x) + \cv(z))^\alpha - \cv(x)^\alpha - \cv(y)^\alpha - \cv(z)^\alpha}{\sum_{w \in \vocab} \cv(w)^\alpha}
\end{align}
Combining the two sides of the equation gives our final result:

\begin{equation}\label{eq:finalrandomdropinequality}
\left( 1 + \frac{c_{\vocab}(x)}{T_{\vocab}} \right)^\alpha > 1 + \frac{ (c_{\vocab}(x) + c_{\vocab}(y))^\alpha + (c_{\vocab}(x) + c_{\vocab}(z))^\alpha - c_{\vocab}(x)^\alpha - c_{\vocab}(y)^\alpha - c_{\vocab}(z)^\alpha}{\sum_{w \in \vocab} c_{\vocab}(w)^\alpha}
\end{equation}

\captionof*{prooff}{Proof of \Cref{thm:randomdropbpe} --- Rényi Entropy increases from \textsc{Random-Drop BPE}.}
\vspace{-1mm}
\end{figure*}

\vspace{-4mm}
\subsection{\textsc{Duplication BPE}}
\label{sec:duplication_bpe}

\citet{zouhar-etal-2023-tokenization} sought to find the Rényi Entropy parameter $\alpha$ that was best correlated with the downstream task performance.
In their setup, the best $\alpha = 2.7$ but $\alpha=3$, the closest integer, is the default in their released toolkit and we use it for all our experiments.
Because this value is greater than 1, it suggests that not only the existence of many very low-frequency tokens hurt performance, but also that having very high-frequency tokens can degrade the performance.

The elimination of rare tokens seems straightforward.
If they are not single-character (atomic), they can be replaced with their constituent parts.
For example \texttt{markup} $\rightarrow$ \texttt{mark -up}.
In some cases this decomposition can increase the frequency of an already-high-frequency token, such as \texttt{again} $\rightarrow$ \texttt{a -gain}, which would increase the frequency of the already common subword \texttt{a}.
In addition, replacing all instances of the high-frequency token will simply add its frequency to two other subwords, making them both appear at least as frequently as the now-eliminated token.
We propose a way to eliminate high-frequency tokens while avoiding these issues.

Given a BPE tokenizer $(\Sigma, \vocab, \merges)$, we select high-frequency tokens and replace them with tokens that have identical surface form (i.e., they spell the same thing) but with different indices (i.e., they have different embeddings in the neural model).
For presentation purposes, we modify the surface form of the token to disambiguate them.
For example, given a high-frequency subword \texttt{-ing}, we denote the duplicates \texttt{-ing}$_1$, \texttt{-ing}$_2$, \ldots, \texttt{-ing}$_k$.

Formally, a \textsc{Duplication BPE} tokenizer is formed by choosing integers $1 \le N \le |\vocab|$ and $2 \le k$. For each of the top $N$ most frequent tokens (measured by their frequency in the corpus), denoted by $\mathcal{X}$, we add $k$ duplicated tokens to $\mathcal{V}$.
Then, during BPE inference, in the final tokenized sequence of a word, if any of the $N$ duplicated subwords appear, we uniformly at random replace it with one of the $k$ duplicates.
\begin{align}
 \sub(z) = \begin{cases} 
    z' \sim \mathcal{U}(\{z_i\}_{1\le i \le k}) & z \in \mathcal{X}   \\
    z & \hspace{-5mm}\text{otherwise}
\end{cases}
\end{align}
Where $\mathcal{U}$ is the discrete uniform distribution (each of the $k$ duplicates have the same probability).
\textsc{Duplication BPE} inference is implemented by replacing Line \ref{algo:inference_return} of Algorithm \ref{algo:bpe_inference} with 
\begin{align}
\textbf{return} \hspace{2mm} \langle \sub(t_i) \mid 1 \leq i \leq |t| \rangle
\end{align}

During the downstream evaluation (e.g., BLEU), we renormalize the tokens by replacing all duplicated tokens with their original surface form (i.e., reverting \texttt{-ing}$_3$ to \texttt{-ing}).
Detokenization and evaluation is then done as normal.
See an example of this modification in \Cref{ex:duplication_bpe}.

\paragraph{Explanation.}
We are eliminating high-frequency subwords by spreading their probability mass across the duplicate tokens.
On the other hand,
the downstream model performance should be bounded above by the baseline due to the fact that the model is not aware that the tokens are duplicates of each other and will learn a distribution that spreads the probability mass between them. In particular, during beam search, the pessimal case is that the cumulative probability mass of the duplicates is higher than any other token, but individually, they are all below the threshold to be included in the beam.

\begin{table}[htbp]
\resizebox{\linewidth}{!}{
\begin{tabular}{
>{\hspace{-1mm}}l<{\hspace{-3mm}}p{6cm}<{\hspace{-1mm}}
}
\toprule
$\pmb{\mathcal{X}}$ & \bf Tokenization \\
\midrule
$\emptyset$ \,\,(original) & \texttt{the apple the berry the citrus the dill the elderberry} \\
$\{\texttt{the}, \texttt{apple}\}$& \texttt{the$_3$ apple$_4$ the$_5$ berry the$_4$ citrus the$_3$ dill the$_7$ elderberry} \\
\bottomrule
\end{tabular}
}
\captionof{myexamplef}{\textsc{Duplication BPE} with $k{=}10$ (second row) changes the token but the surface form remains the same (\texttt{apple$_4$} = \texttt{apple$_3$}).}
\label{ex:duplication_bpe}
\vspace{-4mm}
\end{table}

\paragraph{Rényi Entropy.}

Formally, we treat the algorithm as modifying repeatedly some element $w_x$ and changing its frequency from $\pdelta(w_x)$ to $\frac{\pdelta(w_x)}{k}$.
It is easy to see that for $k=1$, the algorithm does not change the unigram distribution.
Therefore, for the following two theorems, assume $k>1$.

\begin{restatable}{theorem}{thmdecomposeshannon}
\label{thm:decomposeshannon}
Let $\pdelta$ be the unigram distribution of a tokenizer and let $\pdelta'$ be the \textsc{Duplication BPE} unigram distribution with $k>1$.
Let $\rvDelta$ and $\rvDelta'$ be the corresponding random variables.
Then $\ent[\rvDelta] < \ent[\rvDelta']$.
\end{restatable}

\begin{figure*}[htbp]
\proofpreamble

Assume the algorithm modifies the frequency of token $w_x$.
\vspace{-3mm}
\begin{align}
& \ent[\rvDelta'] = -\sum_{w\in \vocab} \pdelta(w)\log(\pdelta) + \pdelta(w_x) \log(\pdelta(w_x)) - k \frac{\pdelta(w_x)}{k}\log\left(\frac{\pdelta(w_x)}{k}\right) \\[-0.5em]
&= \ent[\rvDelta] + \pdelta(w_x)\left(\log(\pdelta(w_x))-\log\left(\frac{\pdelta(w_x)}{k}\right)\right) = \ent[\rvDelta] + \pdelta(w_x) \log(k) > \ent[\rvDelta]
\end{align}
\vspace{-5mm}
\captionof*{prooff}{Proof of \Cref{thm:decomposeshannon} --- Shannon Entropy increases from \textsc{Duplication BPE}.}
\vspace{3mm}
\end{figure*}

\begin{restatable}{theorem}{thmdecomposerenyi}
\label{thm:decomposerenyi}
Let $\pdelta$ be the unigram distribution of a tokenizer and let $\pdelta'$ be the \textsc{Duplication BPE} unigram distribution with $k>1$.
Let $\rvDelta$ and $\rvDelta'$ be the corresponding random variables.
Then,\ $\ent_\alpha[\rvDelta] < \ent_\alpha[\rvDelta']$.
\end{restatable}

\begin{figure*}[htbp]
\proofpreamble

Assume the algorithm modifies the frequency of token $w_x$.
For $\alpha > 1$: $\ent_\alpha[\rvDelta] < \ent_\alpha[\rvDelta'] $.
\begin{align}
\frac{1}{1-\alpha}\log \left( \sum_{w \in \vocab} \pdelta(w)^{\alpha} \right)
&< \frac{1}{1-\alpha}\log \left( \sum_{w \in \vocab}\pdelta(w)^{\alpha} - \pdelta(w_x)^{\alpha} + k\cdot \left(\frac{\pdelta(w_x)}{k}\right)^\alpha \right) \hspace{-1.5cm} \\
\log \left( \sum_{w \in \vocab}\pdelta(w)^{\alpha} \right) 
&\pmb{\pmb{>}} \log \left( \sum_{w \in \vocab} \pdelta(w)^{\alpha}- \pdelta(w_x)^{\alpha} + k\cdot \left(\frac{\pdelta(w_x)}{k}\right)^\alpha \right) \hspace{-8mm}
&\prooftext{(from $\alpha > 1$)} \\
\sum_{w \in \vocab}\pdelta(w)^{\alpha}
&> \sum_{w \in \vocab} \pdelta(w)^{\alpha} - \pdelta(w_x)^{\alpha} + k\cdot \left(\frac{\pdelta(w_x)}{k}\right)^\alpha \\[-0.6em]
0 &> -\pdelta(w_x)^{\alpha} + k\cdot \left(\frac{\pdelta(w_x)}{k}\right)^\alpha \\
0 &> \pdelta(w_x)^{\alpha} \cdot (k^{1-\alpha} - 1) \\
1 &> k^{1-\alpha} & \prooftext{\hspace{-5cm}(holds from $\alpha >1$ and $k>1$)}
\end{align}

The proof follows by reversing the order of equivalent inequalities.
Conversely, for $\alpha < 1$: $\ent_\alpha[\rvDelta] < \ent_\alpha[\rvDelta']$.
\begin{align}
\frac{1}{1-\alpha}\log \left( \sum_{w \in \vocab} \pdelta(w)^{\alpha} \right) &< \frac{1}{1-\alpha}\log \left( \sum_{w \in \vocab}\pdelta(w)^{\alpha} - \pdelta(w_x)^{\alpha} + k\cdot \left(\frac{\pdelta(w_x)}{k}\right)^\alpha \right) \hspace{-1.5cm} \\
\log \left( \sum_{w \in \vocab}\pdelta(w)^{\alpha} \right) &< \log \left( \sum_{w \in \vocab} \pdelta(w)^{\alpha}- \pdelta(w_x)^{\alpha} + k\cdot \left(\frac{\pdelta(w_x)}{k}\right)^\alpha \right) \hspace{-8mm}&\prooftext{(from $\alpha < 1$)} \\
\sum_{w \in \vocab}\pdelta(w)^{\alpha} &< \sum_{w \in \vocab} \pdelta(w)^{\alpha} - \pdelta(w_x)^{\alpha} + k\cdot \left(\frac{\pdelta(w_x)}{k}\right)^\alpha \\[-0.6em]
0 &< -\pdelta(w_x)^{\alpha} + k\cdot \left(\frac{\pdelta(w_x)}{k}\right)^\alpha \\
0 &< \pdelta(w_x)^{\alpha} \cdot (k^{1-\alpha} - 1) \\
1 &< k^{1-\alpha} & \prooftext{\hspace{-5cm}(holds from $\alpha <1$ and $k>1$)}
\end{align}
The proof follows by reversing the order of equivalent inequalities.
For $\alpha=1$ the \Cref{thm:decomposeshannon} applies.
This covers all cases of admissible $\alpha>0$ and therefore concludes the proof.

\captionof*{prooff}{Proof of \Cref{thm:decomposerenyi} --- Rényi Entropy increases from \textsc{Duplication BPE}.}
\vspace{1mm}
\end{figure*}

An illustration of how one application affects the Shannon and Rényi entropies is in \Cref{ex:reduplication_entropy}.

\newcommand{\ulB}[1]{{
    \hspace{-1mm}\setul{0.2ex}{0.3ex}\setulcolor{Brown3}\ul{#1}}\hspace{-1mm}
}
\newcommand{\ulC}[1]{{
    \hspace{-0.8mm}\setul{0.2ex}{0.3ex}\setulcolor{SpringGreen4}\ul{#1}}\hspace{-0.8mm}
}
\newcommand{\ulD}[1]{{
    \hspace{-1mm}\setul{0.2ex}{0.3ex}\setulcolor{Snow4}\ul{#1}}\hspace{-1mm}
}

\begin{table}[htbp]
\renewcommand{\arraystretch}{1.3}
\resizebox{\linewidth}{!}{
\begin{tabular}{
>{\hspace{-2mm}}c<{\hspace{-3mm}}
l<{\hspace{-3mm}}
c<{\hspace{-3mm}}
c<{\hspace{-3mm}}
c<{\hspace{-3mm}}
c<{\hspace{-3mm}}
c<{\hspace{-3mm}}
c<{\hspace{-2mm}}
}
\toprule
$\boldsymbol{k}$ & \bf Unigram Distribution & $\boldsymbol{\ent}$ & $\boldsymbol{\ent_{0.5}}$ & $\boldsymbol{\ent_{3}}$ & $\boldsymbol{\eff}$ & $\boldsymbol{\eff_{0.5}}$ & $\boldsymbol{\eff_{3}}$ \\
\midrule
- & $\langle$\hspace{-1.5mm} \ulD{$0.4$,} $0.3$, $0.2$, $0.1$ \hspace{-1.2mm}$\rangle$ & 1.85 & 1.92 & 1.66 & 1.33 & 1.38 & 1.20 \\
$2$ & $\langle$\hspace{-1.5mm} \ulC{$0.2$, $0.2$,} $0.3$, $0.2$, $0.1$\hspace{-1.2mm} $\rangle$ & 2.25 & 2.28 & 2.13 & 1.40 & 1.42 & 1.33  \\
$10$ & $\langle$\hspace{-1.5mm} \ulC{$0.04$, ..., $0.04$,} $0.3$, ... \hspace{-1.2mm} $\rangle$  & 3.18 & 3.45 & 2.39   & \ulB{1.24} & \ulB{1.35} & \ulB{0.93} \\
\bottomrule
\end{tabular}
}
\captionof{myexamplef}{Entropy/efficiency change after applying \textsc{Duplication BPE} to the \ulD{first element} in the original sequence (first row) which is expanded to \ulC{two} or \ulC{ten} elements. Note the \ulB{case} where efficiency decreases, despite entropy increase, which is an artifact of the small initial vocabulary size.}
\label{ex:reduplication_entropy}
\vspace{-3mm}
\end{table}

\paragraph{Rényi Efficiency.}

We do not have a formal proof that this algorithm increases efficiency because of the change to the denominator $\log|\vocab|$ in \Cref{eq:Rényi_entropy}.
In fact, we show a counterexample where this algorithm \textit{decreases} the efficiency in \Cref{ex:reduplication_entropy}.
When the duplication factor $k$ is high and the vocabulary small, the size of the resulting vocabulary grows too quickly.
Even if $\boldsymbol{\ent}$ grows (as it does in every example), $\log(|\vocab'|) = \log(|\vocab| + (k-1)N)$ grows faster and $\boldsymbol{\eff}$ decreases.
For realistic vocabulary sizes, $\geq 4$k, the relative increase in vocabulary size is much smaller and so $\boldsymbol{\ent}$ increasing tends to imply that $\boldsymbol{\eff}$ will also increase.
We later (\Cref{tbl:duplication_results,tbl:random_drop_results}) show empirically that for larger vocabularies the efficiency increases.

\begin{table}[htbp]
\centering
\renewcommand{\arraystretch}{1.1}
\resizebox{\linewidth}{!}{%
\begin{tabular}{
>{\hspace{-2mm}}c<{\hspace{-4mm}}
c<{\hspace{-3mm}}
c<{\hspace{-2mm}}
c<{\hspace{-2mm}}
c<{\hspace{-2mm}}
c<{\hspace{-2mm}}
c<{\hspace{-2mm}}
}
\toprule
&&&\multicolumn{2}{c}{\bf Overall} &\multicolumn{2}{c}{\bf Best} \\
\bf Tokenizer & $\boldsymbol N$ & $\boldsymbol k$ & $\boldsymbol{\eff_\alpha}$ & \bf BLEU &  $\boldsymbol{\eff_\alpha}$* & \bf BLEU* \\
\midrule
\textsc{Baseline (4k/4k)} & - & - & $0.474$ & $33.74$ & - & - \\ \hdashline
\multirow{ 4}{*}{\shortstack[c]{\textsc{Random-Drop}\\ (4k/4k)}} 
& 2k & 500 & $0.500$ & $33.39$ & $0.504$ & $33.48$ \\
& 2k & 1k & $0.474$ & $32.76$ & $0.483$ & $32.89$ \\
& 4k & 500 & $0.497$ & $33.72$ & $0.498$ & $33.85$ \\
& 4k & 1k & $0.506$ & $33.40$ & $0.518$ & $33.48$ \\ \hdashline
\multirow{ 2}{*}{\shortstack[c]{\textsc{Random-Drop}\\ (4.5k/4.5k)}} 
& 2k & 500 & $0.491$ & $33.35$ & $0.495$ & $33.37$ \\
& 4.5k & 500 & $0.485$ & $33.69$ & $0.487$ & $33.81$ \\
\midrule
\textsc{Baseline (6k/6k)} & - & - & $0.444$ & $33.94$ & - & - \\ \hdashline
\multirow{ 4}{*}{\shortstack[c]{\textsc{Random-Drop}\\ (6k/6k)}} 
& 2k & 500 & $0.468$ & $33.46$ & $0.471$ & $33.46$ \\
& 2k & 1k & $0.441$ & $32.86$ & $0.445$ & $33.03$ \\
& 6k & 500 & $0.458$ & $33.69$ & $0.458$ & $33.94$ \\
& 6k & 1k & $0.473$ & $33.60$ & $0.472$ & $33.71$ \\ \hdashline
\multirow{ 2}{*}{\shortstack[c]{\textsc{Random-Drop}\\ (6.5k/6.5k)}} 
& 2k & 500 & $0.462$ & $33.37$ & $0.464$ & $33.44$ \\
& 6.5k & 500 & $0.451$ & $33.69$ & $0.453$ & $33.70$ \\

\midrule
\textsc{Baseline (9k/9k)} & - & - & $0.418$ & $33.60$ & - & - \\ \hdashline
\multirow{ 4}{*}{\shortstack[c]{\textsc{Random-Drop}\\ (9k/9k)}}
& 2k & 500 & $0.441$ & $33.35$ & $0.440$ & $33.45$ \\
& 2k & 1k & $0.426$ & $32.73$ & $0.440$ & $32.87$ \\
& 9k & 500 & $0.426$ & $33.68$ & $0.425$ & $33.74$ \\
& 9k & 1k & $0.435$ & $33.59$ & $0.436$ & $33.66$ \\ \hdashline
\multirow{ 2}{*}{\shortstack[c]{\textsc{Random-Drop}\\ (9.5k/9.5k)}} 
& 2k & 500   & $0.440$ & $33.31$ & $0.440$ & $33.33$ \\
& 9.5k & 500 & $0.423$ & $33.60$ & $0.423$ & $33.71$ \\
\midrule
\textsc{Baseline (14k/14k)} & - & - & $0.394$ & $33.59$ & - & - \\ \hdashline
\multirow{ 4}{*}{\shortstack[c]{\textsc{Random-Drop}\\ (14k/14k)}} 
& 2k & 500  & $0.415$ & $33.05$ & $0.417$ & $33.09$ \\
& 2k & 1k   & $0.409$ & $32.63$ & $0.416$ & $32.84$ \\
& 14k & 500 & $0.399$ & $33.41$ & $0.399$ & $33.54$ \\
& 14k & 1k  & $0.405$ & $33.35$ & $0.404$ & $33.45$ \\ \hdashline
\multirow{ 2}{*}{\shortstack[c]{\textsc{Random-Drop}\\ (14.5k/14.5k)}} 
& 2k    & 500 & $0.412$ & $33.11$ & $0.415$ & $33.17$ \\
& 14.5k & 500 & $0.396$ & $33.49$ & $0.396$ & $33.54$ \\
\bottomrule
\end{tabular}}
\caption{Experimental results for the \textsc{Random-Drop}. For each \textsc{Baseline} tokenizer, we compare to \textsc{Random-Drop} (based on the respective \textsc{Baseline}) with various hyperparameters. \textbf{Overall} and \textbf{Best} are based on three tokenizer seeds.
}
\label{tbl:random_drop_results}
\end{table}

\begin{table}[htbp]
\centering
\renewcommand{\arraystretch}{1.1}
\resizebox{0.75\linewidth}{!}{%
\begin{tabular}{
c<{\hspace{-3mm}}
l<{\hspace{-3mm}}
c<{\hspace{-2mm}}
c<{\hspace{-2mm}}
c<{\hspace{-2mm}}
}
\toprule
\bf Tokenizer & $\boldsymbol N$ & $\boldsymbol k$ & $\boldsymbol{\eff_\alpha}$ & \bf BLEU \\
\midrule
\textsc{Baseline (4k/4k)} & - & - & $0.474$ & $33.74$ \\ \hdashline
\multirow{ 4}{*}{\shortstack[c]{\textsc{Duplication}\\ \textsc{(4k/4k)}}} 
& $100$ & $3$ & $0.594$ & $32.37$  \\
& $100$ & $5$ & $0.648$ & $31.32$ \\
& $500$ & $3$ & $0.583$ & $32.26$ \\
& $500$ & $5$ & $0.627$ & N/A \\
\midrule
\textsc{Baseline (6k/6k)~~~~~} & - & - & $0.444$ & $33.94$ \\ \hdashline
\multirow{ 4}{*}{\shortstack[c]{\textsc{Duplication}\\ \textsc{(6k/6k)}}} 
& $100$ & $3$ & $0.560$ & $32.27$ \\
& $100$ & $5$ & $0.612$ & $31.60$ \\
& $500$ & $3$ & $0.552$ & $32.43$ \\
& $500$ & $5$ & $0.598$ & $30.57$ \\
\midrule
\textsc{Baseline (9k/9k)} & - & - & $0.418$ & $33.60$ \\ \hdashline
\multirow{ 4}{*}{\shortstack[c]{\textsc{Duplication}\\ \textsc{(9k/9k)}}} 
& $100$ & $3$ & $0.530$ & $32.48$  \\
& $100$ & $5$ & $0.581$ & $31.64$ \\
& $500$ & $3$ & $0.525$ & $32.42$ \\
& $500$ & $5$ & $0.572$ & $30.93$ \\
\midrule
\textsc{Baseline (14k/14k)} & - & - & $0.394$ & $33.59$ \\ \hdashline
\multirow{ 4}{*}{\shortstack[c]{\textsc{Duplication}\\ \textsc{(14k/14k)}}} 
& $100$ & $3$ & $0.501$ & $32.49$ \\
& $100$ & $5$ & $0.551$ & $31.45$ \\
& $500$ & $3$ & $0.498$ & $32.55$ \\
& $500$ & $5$ & $0.545$ & $30.70$ \\
\bottomrule
\end{tabular}
}
\caption{Experimental results for the \textsc{Duplication}. For each \textsc{Baseline} tokenizer, we compare to \textsc{Duplication} (based on the respective \textsc{Baseline}) with various hyperparameters. The average over three training runs is used for all experiments.
N/A: This configuration achieved less than 5 BLEU across several trials, so it is omitted.
}
\label{tbl:duplication_results}
\end{table} 

\vspace{-3mm}
\section{Experiments}
\label{sec:experiments}

We use the same MT Transformer model as \citet{zouhar-etal-2023-tokenization}, \texttt{transformer-iwslt} in fairseq, and the \texttt{iwslt14} German$\rightarrow$English corpus for training.\footnote{This is different than the dataset used by \citet{zouhar-etal-2023-tokenization}, but we chose to use a smaller corpus due to the computational costs.}

We do not use a joint vocabulary between the source and target languages so that we can tune them both independently.
Through hyperparameter search, we found that the 6k/6k BPE model performed the best among all BPE vocabulary sizes (even with different source and target sizes), and we use it as a baseline.
We also experiment with 4k/4k, 9k/9k, and 14k/14k baselines to demonstrate that our results hold across a variety of hyperparameters and vocabulary sizes.
For both of our proposed tokenizer families, we use the baseline BPE tokenizers as the starting point.

All reported predictor metrics were computed using \texttt{tokenization-scorer}\footnote{\href{https://github.com/zouharvi/tokenization-scorer}{github.com/zouharvi/tokenization-scorer}} in their default settings and on the concatenated source+target tokenized training corpora. In particular, we use $\alpha = 3$ for all Rényi efficiency metrics.

\subsection{\textsc{Random-Drop BPE}}

For this family, for each baseline BPE tokenizer (each of the baseline vocabulary sizes), we experimented with four hyperparameter settings. We varied the range of tokens that were candidates for being chosen for decomposition ($N$) with two settings: 2000 and $|\mathcal{V}|$. We also varied the number of tokens that were chosen to be marked for decomposition ($k$) with two settings: $500$ and $1000$.
For each configuration, we created three models using different seeds to choose $\mathcal{D}$. These three models have the same vocabulary size, but have different $\mathcal{D}$ and therefore produce different tokenizations.
We train each configuration with three different seeds and take take the average BLEU score (of all nine models) as the representative score (\textbf{Overall} in \Cref{tbl:random_drop_results}).
To mimic a more real-life setup, we also include the models with the best average scores for each configuration (\textbf{Best} in \Cref{tbl:random_drop_results}). 

\Cref{tbl:random_drop_results} contains the experimental results for this family.
Across nearly every experiment, the \textsc{Random-Drop BPE} tokenizers increase Rényi efficiency compared to their baseline BPE tokenizers \textit{and} the average BLEU scores are higher. In one case, (6k/6k, $N=2000$, $k=1000$), the overall Rényi efficiency is lower than the baseline. However, the efficiency difference is very slight (a -0.68\% relative difference), while the corresponding drop in BLEU is still very large (-1.08 BLEU). In four cases, the best performing model in a hyperparameter setting outperformed the \textsc{Baseline}, but in all of these cases, it was when all tokens were available to be marked for decomposition.
It is likely that infrequent tokens which are unimpactful were selected for decomposition, and so the model performance was not negatively impacted.
When restricted to only the top 2000 most frequent tokens, all models underperform the baseline.

To demonstrate that this finding is not caused purely by the smaller vocabulary size, we also experimented with setting $k = 500$ and using a new vocab size of $|\vocab|+500$ for each \textsc{Baseline}.
In these cases, we observe the same negative correlation between Rényi efficiency and BLEU, indicating that the effective vocabulary size is not the salient factor.

In every case, the \textsc{Random-Drop BPE} models had higher Rényi efficiency than the baseline but performed worse.
Therefore, this provides a counterexample to \citet{zouhar-etal-2023-tokenization}.

\subsection{\textsc{Duplication BPE}}
Again, we vary the hyperparameters to form different \textsc{Duplication BPE} models.
In particular, we set $N$ (the top frequent tokens to duplicate) to $100$ or $500$ and $k$ (the duplication factor) to $3$ or $5$.
We train each configuration with three seeds and report the averages (see \Cref{tbl:duplication_results}).\footnote{We do not report a \textbf{Best} for \textsc{Duplication BPE}, as the choice of $N$ and $k$ represents a \textit{specific} tokenizer. For \textsc{Random-Drop BPE}, the choice of $N$ and $k$ represent a \textit{family} of tokenizers (as $\mathcal{D}$ is selected at random each time, based on those hyperparameters).}

In all cases, the \textsc{Duplication BPE} tokenizer improves Rényi efficiency over the baseline and dramatically reduces BLEU (by at least 1 BLEU).
The largest drops in BLEU --- and simultaneously the largest gains in efficiency --- are with a larger duplication factor: $k=5$.
In the pessimal case (4k/4k, $N = 500$, $k=5$), the model was unable to converge (across 10 trials, none had higher than 5 BLEU).
We conjecture that this is due to the large percentage of tokens ($\sfrac{1}{8}$ of the total vocabulary size) being duplicated $5$ times, so the model is essentially forced to spread the marginal probability of a subword sequence over $|S|^5$ identical-surface-form sequences (where $|S|$ is the sequence length).
In other words, while the duplicated tokens combined might have the largest probability, this probability is divided by 5 which prevents the beam search from selecting any of the correct tokens.

In all cases, we increased the Rényi efficiency while decreasing BLEU, providing yet another counterexample to \citet{zouhar-etal-2023-tokenization}.

\subsection{Other Metrics}
\citet{zouhar-etal-2023-tokenization} considered several other intrinsic metrics, introduced in \Cref{sec:renyi_efficiency}, before determining that Rényi efficiency was the best correlated with downstream performance.
In \Cref{tab:all_metrics}, we list all of these metrics to show which properly correlate with the downstream performance.

For \textsc{Random-Drop BPE}, in all cases both \textbf{PCT} and \textbf{SEQ} correctly predict that the BPE variant will be worse than the baseline.
This means that \textbf{PCT} is always lower than the baseline, \textbf{SEQ} is always higher than the baseline, and BLEU is always worse.
Thus, these non-entropy-based predictors succeed where Rényi efficiency fails.

On the other hand, both \textbf{PCT} and \textbf{SEQ} fail to accurately predict the performance of \textsc{Duplication BPE} models.
For \textbf{SEQ}, the \textsc{Duplication BPE} models always have the same value as the baseline.
This is because these models do not change the surface form of the tokenization, and so, the sequence lengths are always the same.
For \textbf{PCT}, the duplication of the most frequent tokens has the effect of replacing the lowest frequency tokens in the percentile range, thus raising the overall probability mass of the range and increasing the metric. 
Since all \textsc{Duplication BPE} models have lower BLEU than the baseline, these metrics fail similarly to Rényi efficiency on this family.

\begin{table}[htbp]
\centering
\renewcommand{\arraystretch}{1.075}
\resizebox{\linewidth}{!}{
\begin{tabular}{
>{\hspace{-1mm}}c<{\hspace{-4mm}}
r
r
c
c
c
}
\toprule

\bf Tokenizer & \hspace{1mm} $\boldsymbol N$ & $\boldsymbol k$ & $\textbf{PCT} \uparrow$ & $\textbf{SEQ} \downarrow$ & \bf BLEU \\
\midrule
\textsc{Base.} (4k/4k)
& - & - & 0.461 & 25.50 & 33.74\\

\hdashline

\multirow{ 4}{*}{\textsc{Random-Drop}}
& 2k & 500 & 0.356 & 31.46 & 33.39\\
& 2k & 1k & 0.233 & 40.37 & 32.76\\
& 4k & 500 & 0.405 & 29.23 & 33.72\\
& 4k & 1k & 0.352 & 33.37 & 33.40\\

\hdashline

\multirow{ 2}{*}{\shortstack[c]{\textsc{Random-Drop} \\ (4.5k/4.5k)}}
& 2k & 500 & 0.356 & 31.46 & 33.35\\
& 4.5k & 500 & 0.402 & 27.93 & 33.69\\

\hdashline

\multirow{ 4}{*}{\textsc{Duplicate}}
& 100 & 3 & 0.590 & 25.50 & 32.37\\
& 100 & 5 & 0.633 & 25.50 & 31.32\\
& 500 & 3 & 0.571 & 25.50 & 32.26\\
& 500 & 5 & 0.605 & 25.50 & N/A\\

\midrule
\textsc{Base.} (6k/6k)
& - & - & 0.405 & 24.05 & 33.94\\

\hdashline

\multirow{ 4}{*}{\textsc{Random-Drop}}
& 2k & 500 & 0.312 & 29.46 & 33.46\\
& 2k & 1k & 0.210 & 37.64 & 32.86\\
& 6k & 500 & 0.383 & 25.79 & 33.69\\
& 6k & 1k & 0.354 & 27.92 & 33.60\\

\hdashline

\multirow{ 2}{*}{\shortstack[c]{\textsc{Random-Drop} \\ (6.5k/6.5k)}}
& 2k & 500 & 0.303 & 29.19 & 33.37\\
& 6.5k & 500 & 0.376 & 25.35 & 33.69\\

\hdashline

\multirow{ 4}{*}{\textsc{Duplicate}} 
& 100 & 3 & 0.525 & 24.05 & 32.27\\
& 100 & 5 & 0.574 & 24.05 & 31.60\\
& 500 & 3 & 0.522 & 24.05 & 32.43\\
& 500 & 5 & 0.562 & 24.05 & 30.57\\

\midrule
\textsc{Base.} (9k/9k)
& - & - & 0.356 & 22.91 & 33.60\\

\hdashline

\multirow{ 4}{*}{\textsc{Random-Drop}}
& 2k & 500 & 0.275 & 28.04 & 33.35\\
& 2k & 1k & 0.190 & 35.56 & 32.73\\
& 9k & 500 & 0.347 & 23.81 & 33.68\\
& 9k & 1k & 0.332 & 25.07 & 33.59\\

\hdashline

\multirow{ 2}{*}{\shortstack[c]{\textsc{Random-Drop} \\ (9.5k/9.5k)}}
& 2k & 500 & 0.273 & 27.70 & 33.31\\
& 9.5k & 500 & 0.339 & 23.83 & 33.60\\

\hdashline

\multirow{ 4}{*}{\textsc{Duplicate}} 
& 100 & 3 & 0.456 & 22.91 & 32.48\\
& 100 & 5 & 0.507 & 22.91 & 31.64\\
& 500 & 3 & 0.469 & 22.91 & 32.42\\
& 500 & 5 & 0.514 & 22.91 & 30.93\\

\midrule
\textsc{Base.} (14k/14k)
& - & - & 0.308 & 21.95 & 33.59\\

\hdashline

\multirow{ 4}{*}{\textsc{Random-Drop}}
& 2k & 500 & 0.244 & 26.61 & 33.05\\
& 2k & 1k & 0.178 & 33.03 & 32.63\\
& 14k & 500 & 0.301 & 22.67 & 33.41\\
& 14k & 1k & 0.295 & 23.33 & 33.35\\

\hdashline

\multirow{ 2}{*}{\shortstack[c]{\textsc{Random-Drop} \\ (14.5k/14.5k)}}
& 2k & 500 & 0.241 & 26.50 & 33.11\\
& 14.5k & 500 & 0.301 & 22.43 & 33.49\\

\hdashline

\multirow{ 4}{*}{\textsc{Duplicate}}
& 100 & 3 & 0.377 & 21.95 & 32.49\\
& 100 & 5 & 0.432 & 21.95 & 31.45\\
& 500 & 3 & 0.411 & 21.95 & 32.55\\
& 500 & 5 & 0.458 & 21.95 & 30.70\\

\bottomrule
\end{tabular}
}
\caption{The Percentile Frequency (\textbf{PCT}) and Sequence Length (\textbf{SEQ}) predictors compared with the BLEU score. \textbf{PCT} positively correlates with BLEU, while \textbf{SEQ} correlates negatively.}
\label{tab:all_metrics}
\end{table}

\section{Conclusion}

We proposed two families of BPE modifications which break the Rényi efficiency hypothesis by \citet{zouhar-etal-2023-tokenization}.
We were able to construct tokenizers such that the Rényi efficiency \textit{negatively} correlates with the downstream performance. Our tokenizer families are designed specifically to increase Rényi efficiency over their baseline tokenizers, and, across all experiments, produce models that have much lower performance on a German$\rightarrow$English translation task. That our results hold across a range of hyperparameter settings indicates that our tokenizer families serve as counterexamples to the Rényi efficiency hypothesis. We also find that the other metrics by \citet{zouhar-etal-2023-tokenization} fail to correctly predict the performance of our synthetic tokenizer families.

\paragraph{Application.}
Despite our work showing counterexamples of where Rényi efficiency fails, it should not be interpreted as a discouragement to use this metric for estimation of tokenization quality.
We show that there exist synthetic counterexamples in the space of \textit{all} possible tokenizations.
However, a practitioner might be deciding between two real tokenizers, based on BPE variants and hyperparameters.
Our counterexample lie outside of the space of commonly used tokenizers.

\paragraph{Related Work.}
In our setup, we assumed a strict separation between the tokenization and model training step.
However, this need not be the case and \cite{xiao-etal-2010-joint,hiraoka2021joint} successfully merge these two steps into one, creating a single pipeline which jointly optimizes the tokenizer and model for the downstream task.
There are also plethora mechanisms by which tokenization problems can be circumvented during training or inference, such as marginalization over possible tokenizations \citep{kudo-2018-subword,he-etal-2020-dynamic,provilkov-etal-2020-bpe}.
These are not currently taken into consideration and their interaction with the hypotheses surrounding predicting NLP model performance based on the tokenization is unknown.
We further acknowledge tokenization-less NLP models which do not necessitate a tokenization metric, though as of 2023 they remain a minority.

\paragraph{Future Work.}

We identified a variance dimension which contradicts the Rényi efficiency hypothesis.
However, the Rényi efficiency tokenization hypothesis itself was created to replace the Rényi entropy tokenization hypothesis because the latter did not take the dimension of vocabulary size into account.
In the same vein, we hope the original hypothesis can be further refined to incorporate the variance from our two counterexamples.

\section*{Bibliographical References}
\bibliography{misc/anthology,misc/custom}
\bibliographystyle{misc/lrec_natbib}

\end{document}